%% file: main.tex
\documentclass[lettersize,journal]{IEEEtran}
\usepackage{amsfonts}
\usepackage{algorithmic}
\usepackage{algorithm}
\usepackage{array}
\usepackage[caption=false,font=normalsize,labelfont=sf,textfont=sf]{subfig}
\usepackage{textcomp}
\usepackage{stfloats}
\usepackage{url}
\usepackage{verbatim}
\usepackage{graphicx}
\usepackage{cite}
\usepackage{booktabs}
\usepackage{multirow}
\usepackage{amsmath}
\usepackage{hyperref}
\allowdisplaybreaks
\usepackage[table,xcdraw]{xcolor}
\usepackage{amssymb}
\usepackage{amsmath}
\usepackage{booktabs}
\usepackage{multirow}
\usepackage{paralist,bm}

\def\eg{\emph{e.g.~}}

\begin{document}

\title{RadioFormer: Pixel-wise Transformer Framework for Fast Radio Map Estimation}
\title{RadioFormer: A Pixel-wise Transformer for Radio Map Estimation with 1\textpertenthousand~Spatial Observation}
\title{RadioFormer: A Multiple-Granularity Radio Map Estimation Transformer with 1\textpertenthousand~Spatial Sampling }

\author{Zheng Fang, Kangjun Liu, Ke Chen,~\IEEEmembership{Member,~IEEE,} Qingyu Liu,~\IEEEmembership{Member,~IEEE,}\\ Jianguo Zhang, Lingyang Song,~\IEEEmembership{Fellow,~IEEE,}~and~Yaowei Wang 

\thanks{This work is supported in part by the Guangdong Pearl River Talent Program (Introduction of Young Talent) under Grant No. 2019QN01X246, the National Natural Science Foundation of China under Grant No. U20B2052 and the Guangdong Basic and Applied Basic Research Foundation under Grant No. 2023A1515011104.}
\thanks{Z. Fang is with the Pengcheng Laboratory, China, and also with the Department of Computer Science and Engineering, Southern University of Science and Technology, Shenzhen, China; 
K. Liu and K. Chen are with the Pengcheng Laboratory, China; 
J. Zhang is with the Department of Computer Science and Engineering, Southern University of Science and Technology, Shenzhen, China;
Q. Liu and L. Song are with the School of Electronic and Computer Engineering, Peking University Shenzhen Graduate School, Shenzhen, China, and also with the Pengcheng Laboratory, China;
Y. Wang is with the Harbin Institute of Technology, Shenzhen, China, and also with the Pengcheng Laboratory, China.}
}

\markboth{Journal of \LaTeX\ Class Files,~Vol.~14, No.~8, August~2021}%
{Shell \MakeLowercase{\textit{et al.}}: A Sample Article Using IEEEtran.cls for IEEE Journals}


\maketitle

\begin{abstract}
The task of radio map estimation aims to generate a dense representation of electromagnetic spectrum quantities, such as the received signal strength at each grid point within a geographic region, based on measurements from a subset of spatially distributed nodes (represented as pixels).
Recently, deep vision models such as the U-Net have been adapted to radio map estimation, whose effectiveness can be guaranteed with sufficient spatial observations (typically 1\textperthousand~$\sim$~1\%~of pixels) in each map, to model local dependency of observed signal power.  
However, such a setting of sufficient measurements can be less practical in real-world scenarios, where extreme sparsity in spatial sampling (\eg 1\textpertenthousand~of pixels) can be widely encountered.
To address this challenge, we propose RadioFormer, a novel multiple-granularity transformer designed to handle the constraints posed by spatial sparse observations. 
Our RadioFormer, through a dual-stream self-attention~(DSA) module, can respectively discover the correlation of pixel-wise observed signal power and also learn patch-wise buildings' geometries in a style of multiple granularities, which are integrated into multi-scale representations of radio maps by a cross-stream cross-attention~(CCA) module.
Extensive experiments on the public RadioMapSeer dataset demonstrate that RadioFormer outperforms state-of-the-art methods in radio map estimation while maintaining the lowest computational cost.  
Furthermore, the proposed approach exhibits exceptional generalization capabilities and robust zero-shot performance, underscoring its potential to advance radio map estimation in a more practical setting with very limited observation nodes.

\end{abstract}


\input{parts/intro}

\input{parts/related_work}

\input{parts/method}

\input{parts/experiment}

\section{Conclusion}
\label{conclusion}

In this paper, we present the RadioFormer, specifically designed to address the challenge of extremely spatial sparse sampling~(\eg 1\textpertenthousand~sampling rate) in radio map estimation. 
Unlike existing vision-based models or the Pixel Transformer concerning feature extraction in a manner of single granularity, RadioFormer independently models observation point features and building features, followed by a fusion mechanism that enables superior prediction accuracy.
Comprehensive experiment results on the public benchmark can demonstrate that our model achieves state-of-the-art performance while maintaining the lowest computational complexity among competing methods. 
Furthermore, our RadioFormer excels in key aspects such as prediction accuracy, zero-shot performance, and generalization capability, underscoring its effectiveness and promising application in real-world scenarios.


\bibliographystyle{IEEEtran}
\bibliography{IEEEabrv,parts/cite}

\end{document}

%% file: parts/intro.tex
\section{introduction}
\begin{figure}[!t]
\centering
\includegraphics[width=3.6 in]{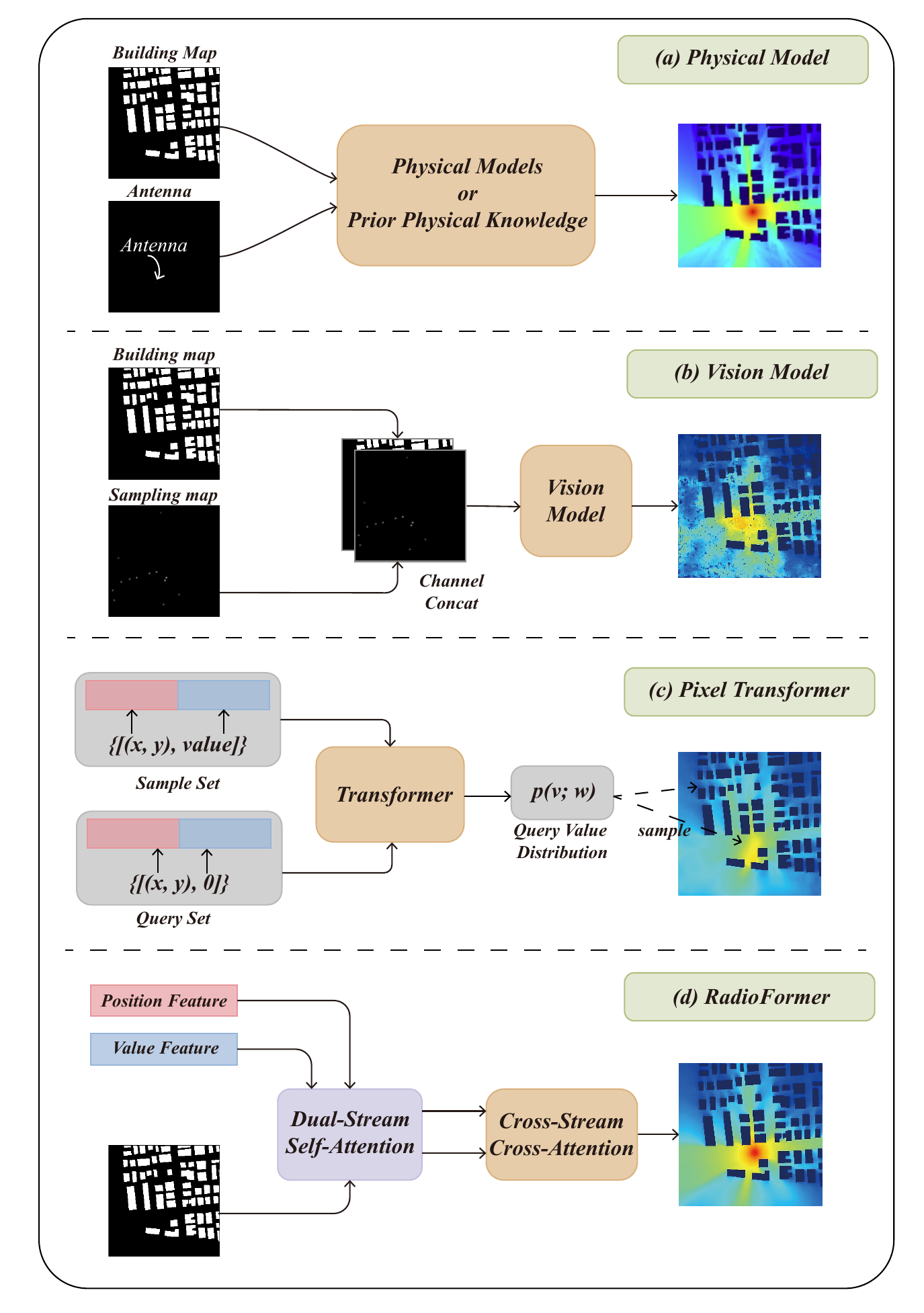}
\caption{Several methods have been proposed for generating common spectrum scenarios: a. \textbf{Physical Method}. This approach leverages physical models or prior knowledge of physical principles to generate a radio map, using building maps and transmitter information as inputs. b. \textbf{Vision Model}. In this method, a sampling map is combined with a building map, and a vision model is then used to predict the distribution of spectrum scenarios. c. \textbf{Pixel Transformer}. This technique generates the spectrum distribution by inferring the values of unknown points based on the known data points and processing them iteratively. d. \textbf{RadioFormer}. A novel approach introduced by our work enhances prediction by incorporating building map information to refine the interactions between known points, resulting in a comprehensive radio map.}
\label{fig_1}
\end{figure}

The electromagnetic spectrum situation encompasses the current state, overarching dynamics, and evolving trends within the electromagnetic environment, forming a foundation for advancing wireless communication systems. As networks demand higher capacity, reliability, and efficiency, understanding and modeling the spectrum has become increasingly critical~\cite{intro1}. In this context, the radio environment map~(REM) plays a pivotal role in quantifying wireless signal degradation between transmitters (Tx) and receivers (Rx), capturing the influence of environmental factors such as propagation loss and building reflections~\cite{intro2}. With the integration of AI into networking, REMs have emerged as indispensable tools for optimizing network performance, interference management, and spectrum utilization~\cite{uav, li2024radio, romero2022radio}. Their applications extend to critical tasks such as communication blind spot detection, resource allocation, and enhancing next-generation AI-driven networking systems~\cite{ckm, intro3, alhammadi2024artificial}.


As shown on the far right of Fig.~\ref{fig_1}~(a), a radio map represents local radio signals, with each pixel indicating signal strength: red denotes the strongest signal, while a gradient from yellow to green to dark blue reflects progressively weaker signals. These pixel values are derived from sensing devices deployed in the environment, which often cover only about $\textpertenthousand$ of the total region to be analyzed. The core objective of spectrum situation generation is to utilize this sparse, localized data to model and characterize the broader electromagnetic spectrum, creating a comprehensive global view of the radio environment. However, this sparse observation sampling presents significant challenges for radio map estimation, particularly in practical applications. Fig.~\ref{fig_1} illustrates several methods for generating radio maps, including physical models or prior knowledge approaches that rely on accurate information about emission sources and building layouts. While these methods are often accurate under controlled conditions, they face limitations in real-world scenarios due to their dependence on precise transmitter locations or dense observation networks, which are rarely available in practice. Additionally, their substantial computational demands restrict their feasibility for real-time or large-scale applications.

In recent years, advancements in deep learning have inspired its application in spectrum situation generation tasks~\cite{deepL2, st_radiomap, 5d_dataset}. As illustrated in Fig.~\ref{fig_1}~(b), some studies~\cite{radiounet, fadenet} have explored the use of vision models for radio map prediction. These approaches encode observation data into a sampling map, which is combined with building maps as two distinct input channels for the model. This strategy has shown promising results in predicting radio maps accurately. However, current methods face significant limitations under extremely sparse observation conditions. Vision models typically rely on a fixed receptive field, processing data within predefined contexts. Sparse sampling leads to severe computational redundancy, increasing the overall computational complexity and diminishing model efficiency and performance.


Pixel-wise deep learning approach~\cite{pixel)_trans} presents a potential solution by reducing redundant computations associated with sparse observations. As shown in Fig.~\ref{fig_1}~(c), these methods employ pixel-by-pixel modeling, with techniques like pixel transformers used to separately model the coordinates and values of observed points. Auto-regressive generative methods~\cite{autoregressive1, autoregressive2} are then used to predict signal distributions, enabling finer granularity in radio map estimation. While this approach addresses redundancy, it introduces considerable computational overhead due to the point-by-point reasoning process. Additionally, these methods fail to incorporate critical environmental factors, such as building layouts, due to the computational burden of including such data at the pixel level. This omission exacerbates challenges under sparse sampling, making radio map estimation an ill-posed problem with no guaranteed unique solution. These dual challenges—high computational cost and the lack of essential contextual information—highlight the need for further model refinement to achieve both efficiency and accuracy in practical applications.


To address these challenges, we propose \textbf{RadioFormer}, a Multi-Granularity Radio Map Estimation Transformer, designed to achieve high model performance while minimizing redundant computations. RadioFormer is specifically tailored for radio map prediction under extremely sparse sampling conditions. By adopting the multi-granularity architecture, it strikes an effective balance between computational efficiency and the need for accurate, detailed predictions in data-limited environments. As depicted in Fig.~\ref{fig_1}~(d), our RadioFormer employs a dual-stream self-attention~(DSA) module, containing two specialized branches, to process inputs at different granularities. A pixel-level branch captures the correlation of pixel-wise observation points, while a patch-level branch processes building geometries. This dual-branch design mitigates redundant complexity in modeling sparse observation data. The multi-granularity features from both branches are subsequently aligned and integrated using a cross-stream cross-attention~(CCA) module, which enables effective interaction between multi-scale information. The integrated features are then decoded into radio maps, bypassing the computationally intensive point-by-point reasoning methods, thereby ensuring both efficiency and precision.


Our key \textbf{contributions} include: 

\begin{itemize}
\item We propose the Multiple-Granularity Transformer for spectrum situation generation, specifically designed to address the task of radio map prediction under conditions of extremely sparse sampling. 
\item We employ a dual-stream self-attention module to process multi-granularity information and a cross-stream cross-attention module to seamlessly integrate pixel-level spectrum features with patch-level spectrum features, thereby mitigating the challenges of long inference times and reducing redundant computational complexity.
\item We conducted abundant experiments using the popular RadioMapSeer dataset, demonstrating that our model achieves the state-of-the-art performance.
\end{itemize}

The remainder of this article is structured as follows: 
Section~\ref{relatework} reviews related work encompassing prior approaches to radio map prediction and advancements in pixel-wise deep learning methods. Section~\ref{preliminaries} formally defines the electromagnetic spectrum situation generation task, providing essential preparatory details, including a concise overview of the transformer architecture and a mathematical formulation of the sampling strategy. 
Section~\ref{method} presents the architecture and workflow of the proposed RadioFormer model. Section V outlines the experimental design, detailing the quantitative and qualitative evaluations conducted to assess model performance. Finally, Section~\ref{conclusion} concludes the article with a summary of findings and implications.
The code will be released at \href{https://github.com/FzJun26th/RadioFormer}{https://github.com/FzJun26th/RadioFormer}.

%% file: parts/related_work.tex
\section{Related Work}
\label{relatework}

\subsection{Radio Map and Pathloss}
A radio map represents the spatial manifestation of pathloss, a key metric in wireless communications used to quantify the reduction in signal strength~(power attenuation) between a transmitter~(Tx) and receiver~(Rx) due to both large-scale effects and small-scale fading. 
According to previous theory\cite{relate_work_pathloss1, relate_work_pathloss2}, the pathloss in dB scale could be defined as:
\begin{equation}
    P_L = (P_{Rx})_{dB} - (P_{Tx})_{dB}, 
\end{equation}
where $P_{Tx}$ and $P_{Rx}$ denote the transmitted power and received power at the Tx and Rx locations, respectively.
To make it suitable for the proposed deep learning estimation method, the pathloss values $P_L$ are converted to gray-level pixel values between 0 and 1 by a function $f$, which is formulated as:
\begin{equation}
f=\max \left\{\frac{\mathrm{P}_{\mathrm{L}}-\mathrm{P}_{\mathrm{L}, \text { trnc }}}{M_1-\mathrm{P}_{\mathrm{L}, \text { trnc }}}, 0\right\},
\end{equation}
where $M_1$ is the maximal pathloss in all radio maps in the dataset and the $P_{\mathrm{L}, \text { trnc }}$ is a physical parameter to truncation and rescale the pathloss function.

\subsection{Radio Map Prediction}
Several methods for estimating radio maps have been proposed in the literature, with a predominant reliance on non-deep learning approaches.  

\subsubsection{Traditional techniques}
Traditional techniques often involve data-driven interpolation~\cite{16, 17}, where the values at observation points are used to reconstruct a complete radio map through various signal-processing methods.   
Other approaches~\cite{21,24} combine observed values with prior knowledge of the physical system, fitting a spatial loss field~(SLF) to estimate radio values at unmeasured locations.   
Additionally, some methods~\cite{25, 26} predict radio maps directly based on available prior knowledge, bypassing the need for explicit measurement data.  
While these methods have demonstrated varying degrees of success, they generally lack the flexibility and scalability of more recent deep learning-based techniques.

\subsubsection{Deep Learning Algorithm}
Deep learning has prompted significant interest in leveraging these techniques for radio map estimation in recent years. Two recent studies~\cite{28, 29} explore using deep learning methods to estimate radio maps. In these approaches, neural networks predict pathloss for each transmitter-receiver~(Tx-Rx) location. The network is trained on a fixed radio map and subsequently used to estimate the radio map at various Tx-Rx locations. This process effectively serves as a data-fitting technique for a four-dimensional~(4D) function, \( G(x, y) \), representing the radio environment. However, one limitation of this approach is that each city map necessitates retraining of the network, with each trained model tailored to the specific characteristics of a given map. This makes the process highly context-dependent and limits the generalizability of the network across different environments.

\subsubsection{Pioneer Architecture}
Radio-UNet~\cite{radiounet} identified key limitations in previous approaches and proposed new requirements and settings to address these issues. First, it emphasized the need for models to learn to approximate the underlying physical phenomena—such as pathloss and signal attenuation—independently of the specific city map, enabling generalization across diverse environments. Second, it advocated for a model input that is agnostic to additional variables, such as the receiver's height or the distance between the receiver and transmitter, thus focusing on the inherent characteristics of the radio environment. 
In response to these challenges, Radio-UNet introduced a novel dataset, RadioSeerMap~\cite{dataset}, designed for training and testing the model. This dataset masks non-observed points and forms a two-channel input graph by concatenating the building map with the sample map. A two-stage, UNet-based architecture is then employed to predict the radio map. This approach facilitates more accurate and generalizable predictions and offers a robust framework for modeling radio propagation in complex urban environments.


\subsection{Pixel-wise Deep Learning Algorithms}
In recent years, pixel-level algorithms have gained attention in image generation tasks.   
PixelCNN~\cite{pixelcnn, pixelcnn++} is one of the earliest methods to generate images pixel by pixel using auto-regressive models. This approach generates the entire image by sequentially predicting each pixel based on the preceding pixels.   
However, the method is inherently inflexible, as the input pixels must follow a fixed order, limiting its ability to model complex dependencies in the image. 
This sequential constraint poses challenges for tasks that require more flexible or parallelized processing. 
Subsequently, Shubham and their colleagues introduced the Pixel Transformer~\cite{pixel)_trans}, a model that predicts an image distribution consistent with observed evidence by incorporating information from randomly selected pixels and their associated color values.   
This approach is particularly intriguing in radio map prediction under extremely sparse sampling conditions, where observation points can be viewed as randomly distributed pixels.   
In contrast to other vision models, the Pixel Transformer is more suited to the radio map prediction task due to its ability to handle sparsely distributed, non-sequential data and effectively model complex dependencies.

An interesting observation made by a paper~\cite{pit} challenges the conventional assumption that inductive biases, such as locality, are essential in modern computer vision architectures.  
The authors found that Transformers could achieve high performance by treating each pixel as a separate token without relying on the locality bias traditionally embedded in convolutional architectures.  
However, while this approach demonstrates promising results, it does not address the high computational cost associated with the quadratic complexity of the Transformer model.  
Thus, although PiT uncovered valuable insights, the issue of computational inefficiency remains unresolved.

Contemporary vision models have shown impressive performance in radio map prediction tasks; however, they struggle in scenarios with extremely sparse observations and are often burdened by significant, redundant computations. Similarly, pixel-wise frameworks are constrained by their high computational complexity, limiting their scalability and efficiency. To overcome these challenges, we propose \textbf{RadioFormer}, a multi-granularity transformer architecture designed to enhance performance under sparse observation conditions while mitigating computational inefficiencies.

%% file: parts/method.tex
\section{Preliminaries}
\label{preliminaries}

\subsection{Problem Definition}
The objective of our task is to simulate radio propagation phenomena $\textit{P}$ to estimate the global radio map $\textbf{M}$ using fewer than 1\textpertenthousand~ observation points $\textbf{O}$ while leveraging the known distribution of buildings $B$. 
The building map \( \textbf{M}_b \) and the radio map \( \textbf{M} \) both have dimensions \( H \times W \). The set of observation points \( \textbf{O} \) consists of \( k \) observation points, where $ k \leq $1 \textpertenthousand~$(H \times W)$. Each observation point \(s_i\) is characterized by two components: its position coordinates \((x_i, y_i)\) and its corresponding signal strength value \( v_i \).

\subsection{Sampling Category}
\label{sc}

This section outlines the methodology employed to simulate observation on the radio map. The building plays a crucial role in radio map estimation tasks, as they are considered opaque and are the primary sources of spectral refraction and reflection. Given this, building maps are typically well-documented. Consequently, we excluded building areas during observation and focused solely on observing non-building regions. The building maps are generally binary, with '0' indicating non-building areas and '1' representing buildings. In practical scenarios, the sampling points are typically uncorrelated, and we adopt \textbf{random sampling $S_r$} to simulate their distribution. The process of random sampling can be mathematically expressed as follows:

\begin{equation}
\begin{aligned}
    S_{r}=& \bigl\{\left[\left(x_i, y_i\right), \textbf{M}\left[x_i, y_i\right]\right] \mid 0 \leq i \leq K-1, i \in N, \\
& 0<x_i<H, 0<y_i<W, \textbf{M}_b\left[x_i, y_i\right] \neq 1\bigl\},
\end{aligned}
\label{f1}
\end{equation}
where $K$ denotes the number of observation points, $\textbf{M}_b$ denotes the building map, $\textbf{M}$ represents the radio map, and $H$ and $W$ correspond to the height and width of the radio map, respectively.

The observation points may be restricted to a specific region in certain specialized scenarios, such as traffic control. We model this using a method called \textbf{constrained sampling $S_c$}. Given the upper and lower bounds, $(H_d, H_u)$ and $(W_d, W_u)$, that define the constrained region, the constrained sampling process can be described by the following procedure:

\begin{equation}
\begin{aligned}
    S_c=& \bigl\{\left[\left(x_i, y_i\right), \textbf{M}\left[x_i, y_i\right]\right] \mid 0 \leq i \leq K-1, i \in N, \\
& H_d<x_i<H_u, W_d<y_i<W_u, \textbf{M}_b\left[x_i, y_i\right] \neq 1 \bigl\}.
\end{aligned}
\label{f2}
\end{equation}

Another special case arises when the sampler is uniformly distributed across the radio map. In this scenario, the region is partitioned into $k \times k$ sub-regions, with one sampler assigned to each area. The bounds of each subregion could be computed by $H_s = H / k$ and $W_s = W / k$.
We employ a method named \textbf{uniform sampling $S_u$} to model this case. The following mathematical procedure can formally express this approach:

\begin{equation}
\begin{aligned}
    S_u=& \bigl\{\left[\left(x_i, y_i\right), \textbf{M}\left[x_i, y_i\right]\right] \mid 0 \leq i \leq k^2 - 1, i \in N, \\
& (i\mid k) \times H_s<x_i< (i \mid k+1 ) \times H_s, \\
& (i \bmod k) \times W_s <y_i<(i \bmod k+1) \times W_s, \\
& \textbf{M}_b\left[x_i, y_i\right] \neq 1 \bigl\},
\end{aligned}
\label{f3}
\end{equation}
where $\mid$ and $\bmod$ represent the mathematical operations division and complement, respectively.

\subsection{Transformer Architecture}
The Transformer model~\cite{transformer} exhibits remarkable capabilities in sequence modeling. Initially proposed for modeling word sequences in natural language processing~(NLP), it has achieved substantial success in that domain~\cite{chatgpt, llama}. Subsequently, the Transformer architecture has been widely adopted across various fields~\cite{vit, llava}, consistently delivering strong performance and yielding favorable results.

The Transformer model is constructed from alternating layers of self-attention and feed-forward networks, with the self-attention mechanism being central to its powerful sequence modeling capabilities. For a given input sequence, the self-attention layer first applies linear transformations to produce three matrices: the Query Matrix (\( \textbf{Q} \)), the Key Matrix (\( \textbf{K} \)), and the Value Matrix (\( \textbf{V} \)). The self-attention layer then computes a weighted sum of the values in \( \textbf{V} \), where the weights are determined by the similarity between the queries \( \textbf{Q} \) and the keys \( \textbf{K} \). This mechanism enables tokens at different positions to interact with each other while preserving the shape of the sequence, such that the output sequence has the same length as the input. The following feed-forward layer performs a series of nonlinear transformations, further enhancing the model's representational capacity and allowing it to capture more complex relationships within the sequence. The following equation could describe the process
of a transformer block $\Phi$ based on self-attention mechanism:

\begin{equation}
\left\{\begin{array}{l}
\textbf{Q} = \textbf{W}_Q \cdot \textbf{X}; \textbf{K} = \textbf{W}_K \cdot \textbf{X}; \textbf{V} = \textbf{W}_V \cdot \textbf{X},  \\
\operatorname{MHSA}\left(\textbf{Q}, \textbf{K}, \textbf{V}\right)=\operatorname{Softmax}(\frac{ \textbf{QK}^{\top}}{\sqrt{d}}\textbf{V}),\\
\tilde\phi(\textbf{X}) = \operatorname{MHSA}\left(\textbf{Q}, \textbf{K}, \textbf{V}\right) + \textbf{X}, \\
\phi(\textbf{X}) = \operatorname{MLP}\left(\tilde\phi(\textbf{X})\right) + \tilde\phi(\textbf{X}).
\end{array}\right.
\end{equation}

However, the self-attention mechanism exhibits a computational complexity of \( O(n^2) \). As the length of the input sequences increases, this can lead to substantial resource demands, potentially causing computational bottlenecks. To address these limitations, we introduce \textbf{RadioFormer} in the next, a novel method that optimizes the attention mechanism, reducing its computational complexity while maintaining its ability to capture multi-granular relationships.
\textbf{\begin{figure*}[!t]
\centering
\includegraphics[height=3.6 in]{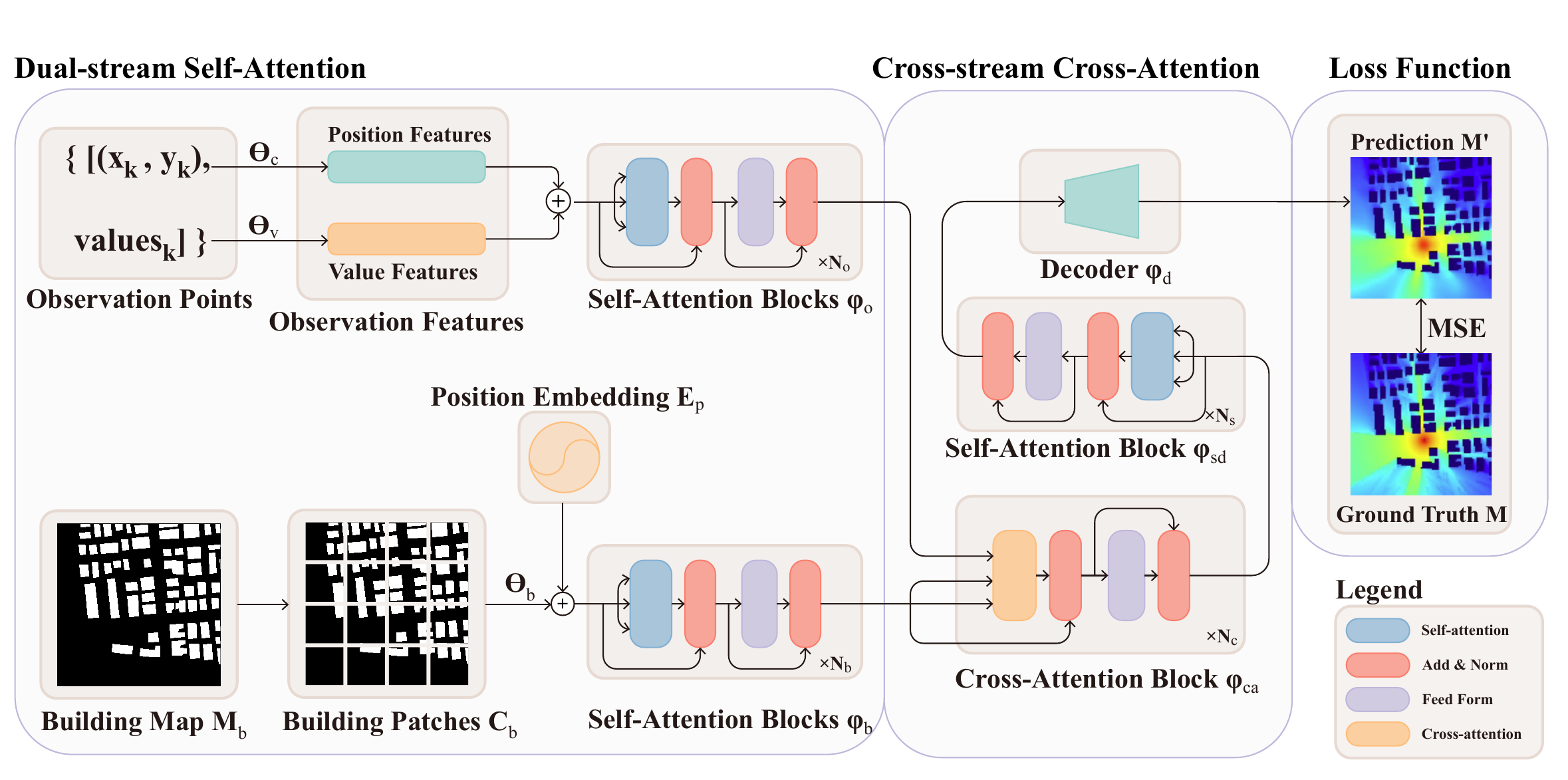}
\caption{\textbf{The workflow for the RadioFormer}: Initially, observation points' position and value information are represented as vectors. These vectors are aggregated to derive the features of the sampling points. Subsequently, self-attention mechanisms facilitate the interaction and fusion of information among the sampling points. This is followed by integrating these fused features with the building map features, which have been encoded using a ViT encoder. This fusion process results in the final prediction map for the situational distribution.}
\label{fig_main}
\end{figure*}}

\section{Method}
\label{method}
This section presents a comprehensive overview of the \textbf{RadioFormer} model architecture, structured into three key components: a multiple granularities encoder: dual-stream self-attention, a feature integration module: cross-stream cross-attention, and the loss function. Each component plays a crucial role in enabling the model to process and integrate multi-granularity information from diverse inputs effectively. The overall workflow of the model, illustrating the sequential interactions between these components, is shown in Fig.~\ref{fig_main}. 
    
\subsection{Dual-stream Self-Attention~(DSA)}  
The \textbf{Dual-stream Self-Attention} (DSA) mechanism employs two distinct encoding branches to process inputs of varying granularities, ensuring that each branch captures information specific to its corresponding level of detail.

The first input to the model is the building map~$\textbf{M}_b$. This map is processed using a standard Vision Transformer (ViT) encoder, which operates at patch-level granularity. This design effectively extracts spatial features and captures the structural details embedded within the building map. Specifically, the building map is divided into multiple patches, forming a set $\textbf{C}_b$. Each patch is then embedded into a vector sequence of dimension $d$ using a linear layer $\theta_b$. Positional embeddings $\textbf{E}_P$ are added to these vectors before they are passed through an encoder consisting of $N_b$ blocks $\phi_b$. The resulting building features $F_B$ are computed as follows:  

\begin{equation}  
\left\{\begin{array}{l}  
\textbf{F}_B^0=\theta_b\left(\textbf{C}_b\right)+\textbf{E}_P \\  
\textbf{F}_B^{(n)}=\phi_b^{(n)}\left(\textbf{F}_B^{(n-1)}\right), \quad n \in \left[1, N_b\right] \\  
\textbf{F}_B = \phi_b^{(N_b)}\left(\textbf{F}_B^{(N_b-1)}\right).  
\end{array}\right.  
\end{equation}  

Each encoder block comprises a Multi-Head Self-Attention (MHSA), a Multilayer Perceptron (MLP) module, a layer normalization (LN), and residual connections. The self-attention mechanism within the MHSA is computed using softmax-weighted interactions between the query, key, and value tokens derived from three learnable linear projection weights.

The second input to the model is a set of observation points, $\textbf{S}$, where each point~$s_i$ is described by its coordinates \((x_i, y_i)\) and an associated value \(v_i\). To encode these inputs, we use two separate linear layers, $\theta_c$ and $\theta_v$, which transform the coordinates and values into \(d\)-dimensional feature vectors referred to as the position and value features, respectively. These two features are combined via element-wise addition to form the initial observation features $\textbf{F}_O^0$, ensuring compatibility with the encoding process for the building features and facilitating the integration of spatial and value information.  

The initial observation features $\textbf{F}_O^0$ are inherently independent and lack explicit inter-point relationships. To enhance the quality of these features, contextual information from the relationships between observation points must be incorporated. For this purpose, a transformer encoder comprising \(N_o\) blocks $\phi_o$ is used to process the observation features $\textbf{F}_O^0$. The final observation features $\textbf{F}_O$ are computed as:  

\begin{equation}  
\left\{\begin{array}{l}  
\textbf{F}_O^0=\theta_c\left(\textbf{S}_C\right)+\theta_v\left(\textbf{S}_V\right) \\  
\textbf{F}_O^{(n)}=\phi_o^{(n)}\left(\textbf{F}_O^{(n-1)}\right), \quad n \in \left[1, N_o\right] \\  
\textbf{F}_O = \phi_o^{(N_o)}\left(\textbf{F}_O^{(N_o-1)}\right).  
\end{array}\right.  
\end{equation}  

This dual-stream architecture effectively encodes both the building map and observation points, capturing multi-granular features essential for subsequent fusion and prediction tasks.

\subsection{Cross-stream Cross-Attention~(CCA)}
\label{cca}

To effectively integrate the building features \(F_B\) and observation features \(F_O\), we design a feature fusion module that addresses two critical considerations. First, the module must effectively combine multi-granularity features, specifically pixel-wise and patch-wise features, to ensure comprehensive feature interaction. Second, the spatial influence of observation points varies based on their relative positions to building patches, necessitating a mechanism to leverage this spatial variation during fusion. Guided by these requirements, we employ cross-attention, a method well-suited for multi-modal feature fusion tasks due to its robust performance. 

Cross-attention~\cite{cross-attention} excels at integrating diverse feature types by combining contextual information from different modalities, producing coherent and logically consistent results. Unlike self-attention, cross-attention operates asymmetrically on two embedded sequences of the same dimensionality. One sequence, denoted as \( \textbf{X} \), serves as the query (\( \textbf{Q} \)), while the other sequence, \( \textbf{Y} \), provides the key (\( \textbf{K} \)) and value (\( \textbf{V} \)). The interaction is formalized as: 

\begin{equation}
\left\{\begin{array}{l}
\textbf{Q} = \textbf{W}_Q \cdot \textbf{X}; \textbf{K} = \textbf{W}_K \cdot \textbf{Y}; \textbf{V} = \textbf{W}_V \cdot \textbf{Y},  \\
\operatorname{CA}\left(\textbf{Q}, \textbf{K}, \textbf{V}\right) = \operatorname{Softmax}\left(\frac{ \textbf{QK}^{\top}}{\sqrt{d}}\right) \cdot \textbf{V}, \\
\tilde{\phi}_\text{CA}(\textbf{X}, \textbf{Y}) = \operatorname{CA}\left(\textbf{Q}, \textbf{K}, \textbf{V}\right) + \textbf{X}, \\
\phi_{CA}(\textbf{X}, \textbf{Y}) = \operatorname{MLP}\left(\tilde{\phi}_\text{CA}(\textbf{X}, \textbf{Y})\right) + \tilde{\phi}_\text{CA}(\textbf{X}, \textbf{Y}).
\end{array}\right.
\end{equation}

This setup allows sequence \( \textbf{Y} \) to focus on \( \textbf{X} \), fostering refined interactions between the two. Additionally, cross-attention enhances model interpretability by explicitly defining the relationships between the sequences.

In our task, where preserving the building map is essential, we treat the building features \(\textbf{F}_B\) as both the key and value, while the observation features \(\textbf{F}_O\) act as the query. The fusion process is carried out using a module composed of \(N_c\) cross-attention blocks, denoted as \(\phi_{ca}\). This module integrates \(\textbf{F}_O\) and \(\textbf{F}_B\) to generate feature map, which is subsequently passed through a module composed of \(N_d\) self-attention blocks, denoted as \(\phi_{sd}\), and a decoder \(\phi_{d}\) to reconstruct the radio map \(\textbf{M}^{\prime}\). This design can be interpreted as leveraging pixel-wise observation features to guide the accurate reconstruction of the radio map based on the building features.

\subsection{Loss Function}
The Pixel Transformer paper suggests that in sample-conditioned signal generation tasks, one should expect the interval distribution for each query point rather than a fixed value to achieve continuous output and predict a multi-modal distribution while mitigating the risk of excessive parameters. 
However, we contend that this approach needs to be better suited for map estimation tasks. 
Unlike color images, radio maps are single-channel, eliminating concerns about the high parameter prediction problem. 
Moreover, the distribution of radio map values is governed by physical laws, and all the values on the radio map are deterministically fixed. Specifically, the radio map is uniquely determined for a given set of sparse observation points and the corresponding building map. Therefore, we argue that Mean Squared Error~(MSE) loss is more appropriate for our network. Therefore, for the output of the CCA module, we employ a lightweight decoder to reconstruct it into the radio map \( \textbf{M}^\prime \). The reconstruction error is then quantified using MSE loss, which measures the discrepancy between the predicted and actual radio map. The formula for calculating MSE loss is as follows:

\begin{equation}
L_\text{MSE}=\frac{1}{N} \sum\left(\textbf{M}-\textbf{M}^{\prime}\right)^2.
\label{f4}
\end{equation}
where \(N\) represents the number of pixels in the radio map. This loss formulation ensures accurate reconstruction of signal distributions and enables effective model optimization.

%% file: parts/experiment.tex
\section{Experiments}
\label{exp}

In this part, we will introduce a series of experiments we designed to validate the effectiveness of our method both qualitatively and quantitatively. 

\subsection{Experiment Setups}
\label{exp_setup}

\textbf{Dataset.} Our experiments utilized the RadioMapSeer dataset~\cite{dataset}, which consists of 700 maps, each with 80 transmitter locations and corresponding coarse analog radio maps. 
These coarse simulations are generated using the dominant path model~(DPM)~\cite{26} and intelligent ray tracing~(IRT)~\cite{ir, ir2} methods, which model two types of interactions between light and geometry. 
The dataset includes several subdatasets, such as DPM and IRT2. 
Additionally, another sub-dataset simulates the first two emitters of each map with higher precision using four interactive, collectively referred to as IRT4. 
All simulations are stored as dense samples of radio maps on a 256$\times256$ meter square grid, where each pixel represents one square meter. 
In these maps, pixel values inside buildings are set to 0, whereas in the building map, the value is 1, with the surrounding area set to 0.

\textbf{Competing Methods.} We compared our model against several widely used vision models, including U-Net~\cite{unet}, CBAM~\cite{cbam}, and Swin-Unet~\cite{swin}. 
These models concatenate the sampling point images and the building maps to form a two-channel input, which is then used to predict the corresponding radio map. 
Furthermore, we evaluated the performance of Pixel Transformer~(PiT)~\cite{pixel)_trans} in generating spectral maps under conditions of extremely sparse sampling. 
Finally, we also compared a state-of-the-art~(SOTA) method in the radio map estimation task, RadioUNet~\cite{radiounet}.
Notably, both methods omit the building map and directly predict the radio map. 

\textbf{Evaluation Protocols.} To assess the performance of our model, we employed three different evaluation metrics. 
The first, Root Mean Square Error~(RMSE)~$\downarrow$, is a widely recognized metric in regression tasks, measuring the average magnitude of the errors between predicted and actual values. 
The second metric, the Structural Similarity Index~(SSIM)~\cite{ssim}, comprehensively evaluates image quality by simulating the human visual system’s perception. 
SSIM~$\uparrow$ considers three key factors: luminance, contrast, and structure, making it a robust indicator of perceptual image fidelity. 
Finally, we used the Peak Signal-to-Noise Ratio~(PSNR)~$\uparrow$, which quantifies the noise level in image reconstruction by comparing the peak signal strength to the noise power, providing a measure of the fidelity of the reconstructed image. 
Together, these three metrics provide a well-rounded evaluation of the model’s performance in radio map estimation tasks.

\textbf{Implement Details.}
In our experiment, we conducted the evaluation using the RadioSeerMap dataset. 
The dataset was partitioned based on the sequence numbers of the building maps. 
Images with sequence numbers larger than 550 were allocated to the test set, while the remaining pictures were randomly split into the training and validation sets in a 12:1 ratio. 
This partitioning strategy ensures that the building maps in the test set are distinct and non-overlapping with those in the training and validation sets, thereby preventing data leakage and ensuring robust evaluation. 
In the RadioFormer model, the dual-stream self-attention module employs two distinct 2-layer self-attention blocks, i.e. \(N_b\) and \(N_o\) are both 2, to process observation points and building maps independently. 
The self-attention block within the building branch operates with a patch size of 16, tailored to capture the structural characteristics of the building maps. 
A single self-attention block facilitates inter-stream information exchange in the cross-stream cross-attention module, i.e. the \(N_c\) and the \(N_s\) are both 1. 
Finally, a convolutional neural network (CNN) decoder is employed as the image reconstruction module, serving as the final stage of the model.
The model was trained using the AdamW optimizer with a Cosine Annealing learning rate scheduler, initializing the learning rate at \(1 \times 10^{-3}\) and the weight decay at \(1 \times 10^{-4}\). 
To mitigate the impact of random sampling on the results, we perform the model evaluation five times and report the average of these runs as the final result.

\begin{figure*}[!t]
\centering
\includegraphics[width= 0.99\textwidth]{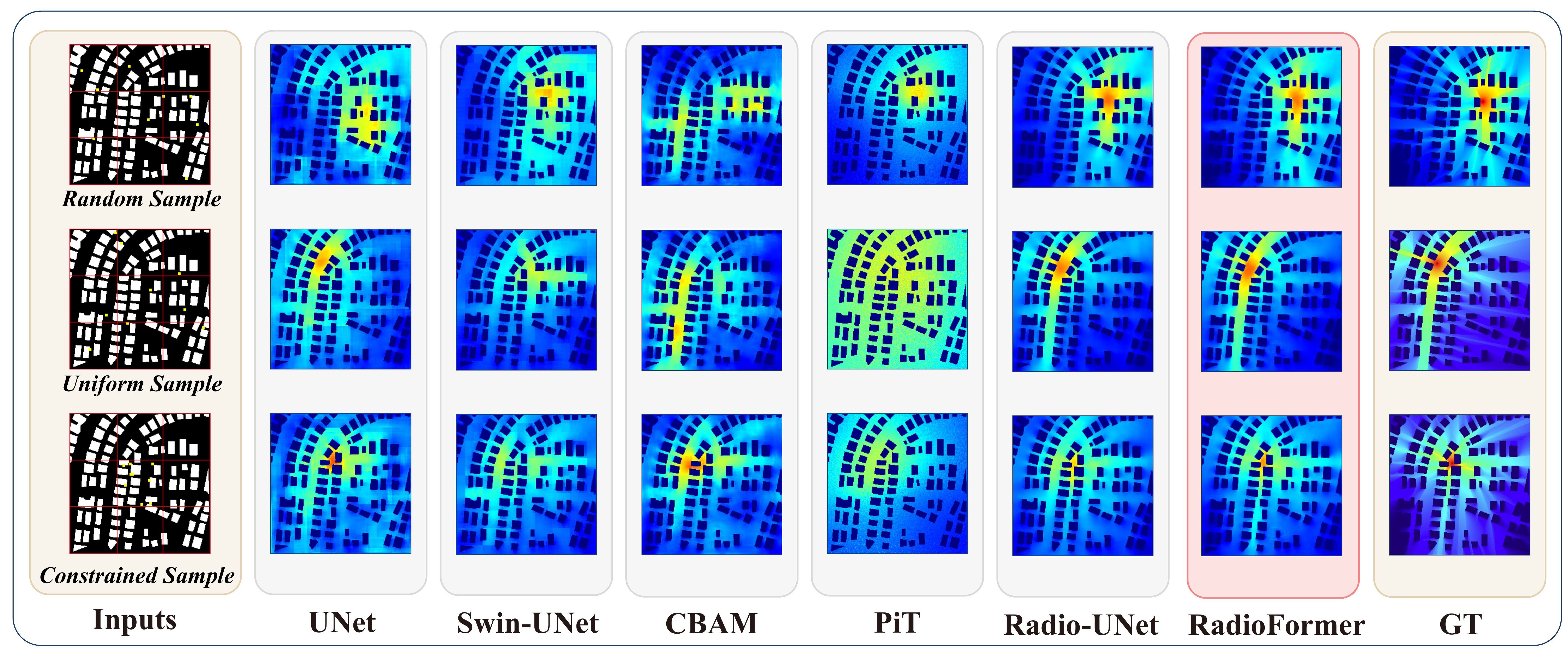}
\caption{Visualization results of various models under 3 different sampling categories. The first column represents three sampling methods. We use yellow dots in these visualizations to indicate the positions of the observation points and red lines to partition the building into distinct regions, thereby highlighting the differences between the three sampling strategies.}
\label{fig3}
\end{figure*}

\begin{table*}[t!]
\caption{Model performance on Different Sampling Categories. The best-performing results are highlighted in bold.}
\centering
\scalebox{0.9}{
\begin{tabular}{@{}l|ccc|ccc|ccc|ccc@{}}
\toprule
\multicolumn{1}{c|}{}                              & \multicolumn{3}{c|}{Random Sample~(Ran)}             & \multicolumn{3}{c|}{Constrained Sample~(Con)}         & \multicolumn{3}{c|}{Uniform Sample~(Uni)}             & \multicolumn{3}{c}{RMSE Fluctuations~$\downarrow$}               \\ \cmidrule(l){2-13} 
\multicolumn{1}{c|}{\multirow{-2}{*}{Dataset:DPM}} & RMSE~$\downarrow$            & SSIM~$\uparrow$            & PSNR~$\uparrow$             & RMSE~$\downarrow$            & SSIM~$\uparrow$            & PSNR~$\uparrow$             & RMSE~$\downarrow$            & SSIM~$\uparrow$            & PSNR~$\uparrow$             & Ran \& Con      & Ran  \& Uni     & Con \& Uni      \\ \midrule
UNet                                               & 0.0736          & 0.8892          & 22.6635          & 0.1076          & 0.8555          & 19.3658          & 0.0817          & 0.8846          & 21.7506          & 0.0340          & 0.0081          & 0.0259          \\
CBAM                                               & 0.0523          & 0.9192          & 25.6307          & 0.0762          & 0.8838          & 24.6198          & 0.0490          & 0.8957          & 26.5789          & 0.0239          & 0.0033          & 0.0272          \\
Swin-UNet                                          & 0.0510          & 0.8927          & 25.8593          & 0.0663          & 0.8926          & 23.5734          & 0.0469          & 0.9001          & 26.5812          & 0.0153          & 0.0041          & 0.0194          \\
Radio-UNet                                         & 0.0421          & 0.9306          & 27.5225          & 0.0520          & 0.9135          & 25.6975          & 0.0388          & 0.9229          & 28.2204          & 0.0099          & 0.0033          & 0.0132          \\
Pixel-Transformer                                  & 0.0820          & 0.8520          & 21.0175          & 0.1032          & 0.7954          & 20.6081          & 0.0849          & 0.8360          & 21.4245          & 0.0212          & 0.0029          & 0.0183          \\
\rowcolor[HTML]{EFEFEF} 
RadioFormer~(ours)                                  & \textbf{0.0338} & \textbf{0.9366} & \textbf{29.4259} & \textbf{0.0416} & \textbf{0.9318} & \textbf{27.6411} & \textbf{0.0319} & \textbf{0.9378} & \textbf{29.9218} & \textbf{0.0078} & \textbf{0.0019} & \textbf{0.0097} \\ \bottomrule
\end{tabular}}
\label{tabel1}
\end{table*}

\begin{figure*}[!t]
\centering
\includegraphics[width= 0.96\textwidth]{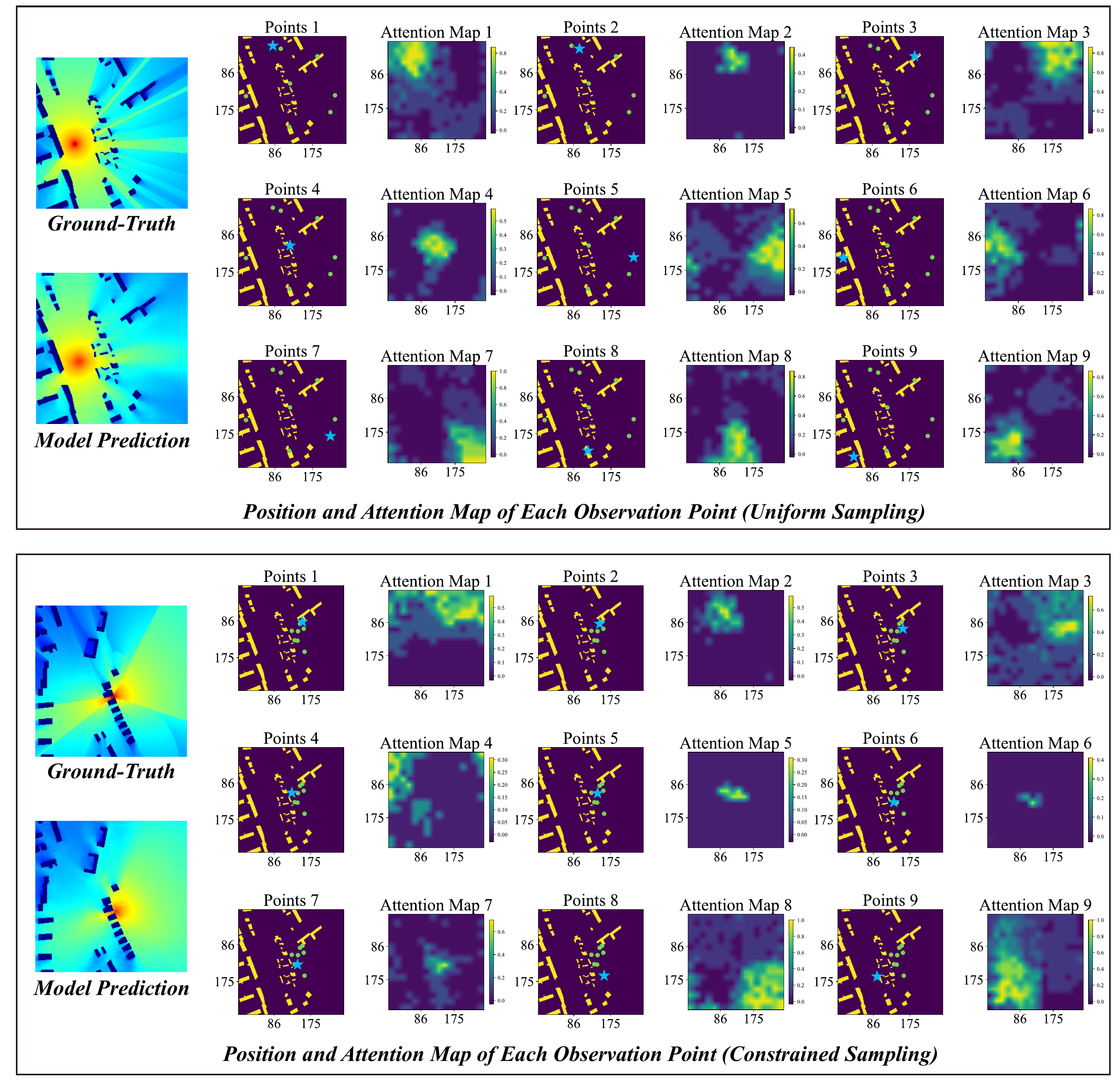}
\caption{Attention map visualization: We visualized the attention map of feature fusion by presenting two images side by side.  The image on the left displays the topographic map, with the positions of the target observation points marked by blue five-pointed stars, while green points mark other observation points. The image on the right illustrates the attention distribution of the observation points for the building map features.  We provide 2 examples with different sampling categories. This visualization provides insight into how the model directs its focus when integrating information from different sources.}
\label{fig_7}
\end{figure*}

\subsection{Comparison With State-of-the-Art Methods}

\textbf{Performance on Different Sampling Method}. 
Fig.~\ref{fig3} presents the visual results of radio map predictions across several different models. 
In each sampling case, the building map is the same, but the locations of the transmitting points vary, which enriches the visualization results. 
The first column of Fig.~\ref{fig3} visualizes three sampling methods—random, constrained, and uniform—described in Section~\ref{sc}. 
The yellow dots in these visualizations indicate the positions of the observation points. 
In the final column, we display the corresponding radio maps, where different colors represent varying signal strengths: red indicates the areas of highest signal strength~(at the transmission point), followed by yellow, green, blue, and black, corresponding to the building areas. 
Tab.~\ref{tabel1} summarizes the performance of each model under the different sampling methods. 

The superiority of our model is evident in Fig.~\ref{fig3}. 
Firstly, our prediction results align with the underlying physical principles. 
For instance, in the constrained sampling scenario, the emission source emits signals predominantly in the central region, with the path directly beneath the source exhibiting strong signal strength~(represented by yellow). 
At the same time, the surrounding areas show weaker signal strength~(depicted in blue). 
Our model's predictions accurately capture this pattern, demonstrating its adherence to the expected physical behavior. 
Secondly, the RadioFormer demonstrates strong stability, as evidenced by the consistent performance of our results across the three sampling methods. 
This stability further suggests that RadioFormer possesses a global receptive field, a characteristic not found in other models. 
The data presented in Tab.~\ref{tabel1} substantiates this claim: our method consistently outperforms others in all scenarios, with the maximum fluctuation in RMSE (among three sample methods) remaining minimal. 
This reinforces the robustness and reliability of the RadioFormer model.

\textbf{Model Explainability.}
We visualized the attention maps of observation points and building features within the CCA~(described in Section~\ref{cca}) module to analyze the model's behavior. 
Fig.~\ref{fig_7} presents attention maps under two distinct sampling conditions, which are uniform and constrained sampling. 
Under uniform sampling, the attention regions~(highlighted in yellow) are relatively consistent across all points, with no notably large or small areas. 
This uniformity indicates that all observation points are effectively utilized. 
In contrast, observation points 1, 8, and 9 under constrained sampling conditions exhibit significantly larger attention regions, while other points, such as points 5,6 and 7, have smaller areas. 
This disparity suggests that points 1, 8, and 9 are over-utilized, whereas others are under-utilized, leading to an imbalance in information contribution.

Further observation reveals that observation points located farther from the region's center are more important in scenarios with dense sampling, while information from points closer to the center is often redundant. 
This observation provides valuable insights for practical applications. It can inform the design of active sampling strategies by imposing constraints to optimize point selection.

\textbf{Performance on Zero-shot and Adaptation.}
The IRT4 sub-dataset is smaller than the DPM sub-dataset and cannot support training the model from scratch. 
However, the IRT4 dataset is characterized by higher precision and sensitivity, making it suitable for evaluating different models' zero-shot and adaptation capabilities. After training the model on an expanded dataset, we evaluated its performance on the IRT4 dataset using both zero-shot and fine-tuning approaches.  Two pre-training scenarios were employed to examine the impact of pre-training data volume on model performance: one utilized the DPM dataset, while the other combined DPM with the IRT2 dataset.  
The data partitioning adhered to the configuration outlined in Section~\ref{exp_setup}, with the number of observation points fixed at 9.
The experimental results, as presented in Tab.~\ref{tabel:fine}, demonstrate the performance of these models under these conditions. 
The results indicate that our model consistently achieves optimal performance across all tested scenarios. A closer examination reveals several key insights. 
First, not all models demonstrate significant zero-shot performance improvements with increased pre-training data; for instance, the UNet model exhibits only marginal gains. 
Second, when the pre-training dataset is DPM, all models benefit from fine-tuning, showing noticeable performance enhancements. 
However, as the size of the pre-training dataset increases, models such as UNet and CBAM struggle to adapt effectively. 
In contrast, our model demonstrates robust adaptability, achieving superior performance under these conditions.
Notably, our model's zero-shot capability consistently exceeds all other models' zero-shot and fine-tuned performance. This result underscores the robustness and effectiveness of our approach, further affirming its superiority in addressing the challenges of diverse experimental conditions.

\begin{table*}[t!]
\caption{Model performance on Zero-shot and Adaptation.}
\centering
\scalebox{0.9}{
\begin{tabular}{@{}l|cccccc|cccccc@{}}
\toprule
\multicolumn{1}{c|}{}                                                                                  & \multicolumn{6}{c|}{Pretrain Dataset: DPM}                                                                                       & \multicolumn{6}{c}{Pretrain Dataset: DPM \& IRT2}                                                                                \\ \cmidrule(l){2-13} 
\multicolumn{1}{c|}{}                                                                                  & \multicolumn{3}{c|}{Zero-shot}                                            & \multicolumn{3}{c|}{Fine-tune}                       & \multicolumn{3}{c|}{Zero-shot}                                            & \multicolumn{3}{c}{Fine-tune}                        \\ \cmidrule(l){2-13} 
\multicolumn{1}{c|}{\multirow{-3}{*}{\begin{tabular}[c]{@{}c@{}}Target Dataset: \\ IRT4\end{tabular}}} & RMSE~$\downarrow$            & SSIM~$\uparrow$            & \multicolumn{1}{c|}{PSNR~$\uparrow$}             & RMSE~$\downarrow$            & SSIM~$\uparrow$            & PSNR~$\uparrow$             & RMSE~$\downarrow$            & SSIM~$\uparrow$            & \multicolumn{1}{c|}{PSNR~$\uparrow$}             & RMSE~$\downarrow$            & SSIM~$\uparrow$            & PSNR~$\uparrow$             \\ \midrule
Swin-UNet                                                                                              & 0.0638          & 0.8611          & \multicolumn{1}{c|}{23.9014}          & 0.0536          & 0.8877          & 25.4134          & 0.0632          & 0.8600          & \multicolumn{1}{c|}{23.9857}          & 0.0559          & 0.8789          & 25.0531          \\
UNet                                                                                                   & 0.0926          & 0.8435          & \multicolumn{1}{c|}{20.6638}          & 0.0669          & 0.8768          & 23.4905          & 0.0730          & 0.8558          & \multicolumn{1}{c|}{22.7323}          & 0.0798          & 0.8425          & 21.9626          \\
CBAM                                                                                                   & 0.0620          & 0.8739          & \multicolumn{1}{c|}{24.1524}          & 0.0510          & 0.9035          & 25.8499          & 0.0582          & 0.8796          & \multicolumn{1}{c|}{24.7014}          & 0.0696          & 0.8643          & 23.1513          \\
Pixel-Transformer                                                                                      & 0.1020          & 0.7520          & \multicolumn{1}{c|}{19.6785}          & 0.0991          & 0.7482          & 20.3321          & 0.9723          & 0.7920          & \multicolumn{1}{c|}{20.6835}          & 0.9700          & 0.7967          & 21.0035          \\
\rowcolor[HTML]{F2F2F2}RadioFormer~(ours)                                                              & \textbf{0.0562} & \textbf{0.8888} & \multicolumn{1}{c|}{\textbf{25.0145}} & \textbf{0.0423} & \textbf{0.9073} & \textbf{26.1823} & \textbf{0.0491} & \textbf{0.9021} & \multicolumn{1}{c|}{\textbf{25.1235}} & \textbf{0.0399} & \textbf{0.9023} & \textbf{26.3265} \\ \bottomrule
\end{tabular}}
\label{tabel:fine}
\end{table*}

\begin{figure*}[!t]
\centering
\includegraphics[width= 0.99\textwidth]{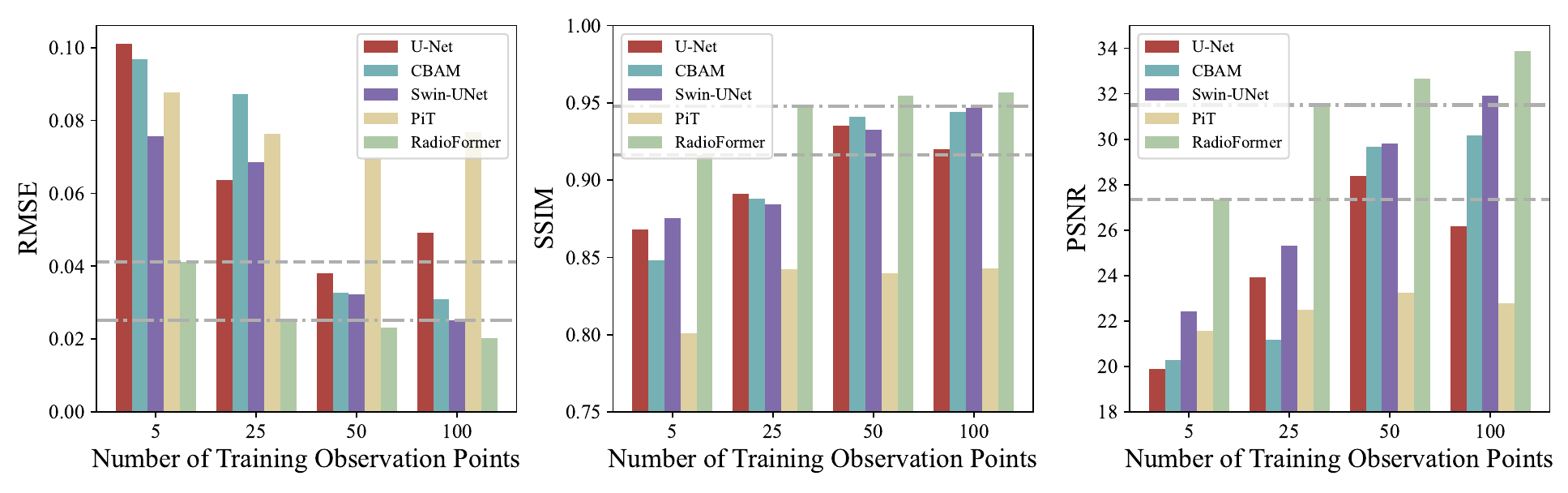}
\caption{The relationship between model performance and the number of observation points during training. Sub-figures 1, 2, and 3 present the RMSE~$\downarrow$, SSIM~$\uparrow$, and PSNR~$\uparrow$ results. In each sub-figure, we have delineated two regions using gray dashed and dotted lines to compare the performance of RadioFormer and other models.}
\label{fig_6}
\end{figure*}

\textbf{Performance on Different Training Observation Points.}
By varying the number of observation points during training, we observed the resulting changes in the evaluation metrics of different models, presented as a bar chart in Fig.~\ref{fig_6}.  
Our model consistently achieves optimal performance regardless of the number of observation points.  
Furthermore, we found that RadioFormer significantly reduces the number of required observation points compared to other methods.  
To illustrate this, we highlighted the performance of RadioFormer when the number of observation points was set to 5 and 25, drawing a gray dashed line and a gray dotted line in each sub-figure of Fig.~\ref{fig_6} to facilitate comparisons across different sampling configurations.   
The results demonstrate several key findings.      
First, the performance of the RadioFormer model trained with only 5 observation points surpasses that of all the pixel transformers.      
Second, the RadioFormer model trained with 5 observation points achieves better performance than all models trained with 25 observation points and even outperforms the UNet model trained with 100 observation points.      
Finally, when comparing the RadioFormer model trained with 25 observation points to UNet, CBAM, and Swin-UNet models trained with 100 observation points, the RadioFormer model significantly outperforms UNet and CBAM, while achieving comparable performance to Swin-UNet.
These findings highlight the RadioFormer model’s ability to fully exploit the information from observation points, demonstrating its efficiency and robustness.     
In practical applications, the RadioFormer model offers the potential to reduce observation point deployment by approximately 75$\%$, providing a substantial advantage in resource-constrained scenarios.

\begin{figure*}[!t]
\centering
\includegraphics[width= 0.99\textwidth]{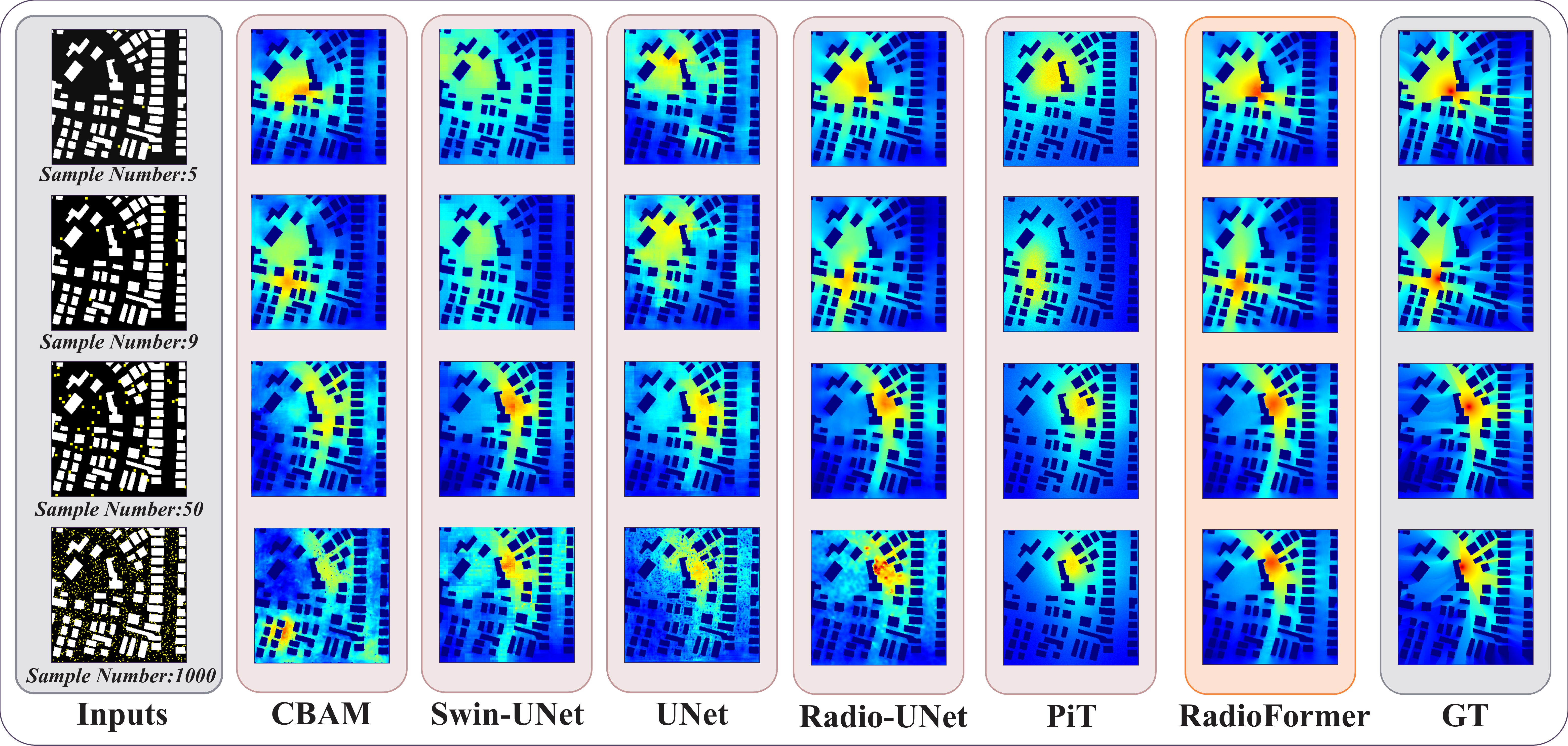}
\caption{The \textbf{visual prediction results} for various models under different observation conditions are presented. The first column displays plots corresponding to different sample sizes or proportions alongside the ground truth. The second through sixth columns show the predicted outputs from different models. The rows correspond to varying numbers of sample points: 5, 9, 100, and 1000, from top to bottom. Notably, our model is highlighted with a red background, while the other models are marked with a yellow background.}
\label{fig_4}
\end{figure*}

\begin{table}[t!]
\caption{Generalization Performance on Different Observation Number.}
\centering
\scalebox{0.8}{
\begin{tabular}{@{}l|ccc|ccc@{}}
\toprule
\multicolumn{1}{c|}{}                                & \multicolumn{3}{c|}{5}                               & \multicolumn{3}{c}{25}                               \\ \cmidrule(l){2-7} 
\multicolumn{1}{c|}{\multirow{-2}{*}{Sample Number}} & RMSE~$\downarrow$            & SSIM~$\uparrow$            & PSNR~$\uparrow$             & RMSE~$\downarrow$            & SSIM~$\uparrow$            & PSNR~$\uparrow$             \\ \midrule
Swin-UNet                                            & 0.0624          & 0.8830          & 24.0935          & 0.0389          & 0.9033          & 28.2035          \\
UNet                                                 & 0.0861          & 0.8777          & 21.2998          & 0.0488          & 0.9173          & 26.2279          \\
CBAM                                                 & 0.0624          & 0.8808          & 24.1004          & 0.0391          & 0.9026          & 28.1501          \\
Radio-UNet                                            & 0.0534          & 0.9122          & 25.4649          & 0.0334          & 0.9261          & 29.5384          \\
Pixel-Transformer                                    & 0.1030          & 0.8532          & 19.7470          & 0.0821          & 0.8679          & 21.7171          \\
\rowcolor[HTML]{F2F2F2} 
RadioFormer~(ours)                                    & \textbf{0.0442} & \textbf{0.9300} & \textbf{27.1230} & \textbf{0.0290} & \textbf{0.9390} & \textbf{30.7665} \\ \midrule
\multicolumn{1}{c|}{}                                & \multicolumn{3}{c|}{50}                              & \multicolumn{3}{c}{75}                               \\ \cmidrule(l){2-7} 
\multicolumn{1}{c|}{\multirow{-2}{*}{Sample Number}} & RMSE~$\downarrow$            & SSIM~$\uparrow$            & PSNR~$\uparrow$             & RMSE~$\downarrow$            & SSIM~$\uparrow$            & PSNR~$\uparrow$             \\ \midrule
Swin-UNet                                            & 0.0349          & 0.9161          & 29.1418          & 0.0355          & 0.9164          & 28.8013          \\
UNet                                                 & 0.0383          & 0.9191          & 28.3287          & 0.0350          & 0.9318          & 29.1119          \\
CBAM                                                 & 0.0357          & 0.9141          & 28.9382          & 0.0357          & 0.9122          & 28.9491          \\
Radio-UNet                                            & 0.0315          & 0.9263          & 30.0369          & 0.0315          & 0.9349          & 30.0366          \\
Pixel-Transformer                                    & 0.0785          & 0.8696          & 22.0987          & 0.0732          & 0.8872          & 22.7568          \\
\rowcolor[HTML]{F2F2F2} 
RadioFormer~(ours)                                    & \textbf{0.0283} & \textbf{0.9384} & \textbf{30.9743} & \textbf{0.0281} & \textbf{0.9380} & \textbf{31.0196} \\ \midrule
\multicolumn{1}{c|}{}                                & \multicolumn{3}{c|}{100}                             & \multicolumn{3}{c}{250}                              \\ \cmidrule(l){2-7} 
\multicolumn{1}{c|}{\multirow{-2}{*}{Sample Number}} & RMSE~$\downarrow$            & SSIM~$\uparrow$            & PSNR~$\uparrow$             & RMSE~$\downarrow$            & SSIM~$\uparrow$            & PSNR~$\uparrow$             \\ \midrule
Swin-UNet                                            & 0.0352          & 0.9261          & 29.6839          & 0.0364          & 0.9196          & 29.7957          \\
UNet                                                 & 0.0366          & 0.9223          & 29.4790          & 0.0358          & 0.9287          & 29.9550          \\
CBAM                                                 & 0.0359          & 0.9203          & 28.9059          & 0.0347          & 0.9244          & 29.1725          \\
Radio-UNet                                            & 0.0320          & 0.9331          & 29.8974          & 0.0363          & 0.9274          & 28.7946          \\
Pixel-Transformer                                    & 0.0758          & 0.8719          & 22.4060          & 0.0739          & 0.8716          & 22.6309          \\
\rowcolor[HTML]{F2F2F2} 
RadioFormer~(ours)                                    & \textbf{0.0281} & \textbf{0.9376} & \textbf{31.0173} & \textbf{0.0282} & \textbf{0.9371} & \textbf{30.9939} \\ \midrule
\multicolumn{1}{c|}{}                                & \multicolumn{3}{c|}{500}                             & \multicolumn{3}{c}{1000}                             \\ \cmidrule(l){2-7} 
\multicolumn{1}{c|}{\multirow{-2}{*}{Sample Number}} & RMSE~$\downarrow$            & SSIM~$\uparrow$            & PSNR~$\uparrow$             & RMSE~$\downarrow$            & SSIM~$\uparrow$            & PSNR~$\uparrow$             \\ \midrule
Swin-UNet                                            & 0.0367          & 0.9026          & 29.1388          & 0.0422          & 0.8574          & 27.4986          \\
UNet                                                 & 0.0358          & 0.9168          & 28.6944          & 0.0488          & 0.8583          & 26.2389          \\
CBAM                                                 & 0.0375          & 0.9105          & 28.5307          & 0.0516          & 0.8715          & 25.7505          \\
Radio-UNet                                            & 0.0434          & 0.8944          & 27.2468          & 0.0578          & 0.8204          & 24.7624          \\
Pixel-Transformer                                    & 0.0731          & 0.8721          & 22.7231          & 0.0735          & 0.8715          & 22.6766          \\
\rowcolor[HTML]{F2F2F2} 
RadioFormer~(ours)                                    & \textbf{0.0283} & \textbf{0.9370} & \textbf{30.9628} & \textbf{0.0284} & \textbf{0.9371} & \textbf{30.9326} \\ \bottomrule
\end{tabular}
}
\label{tabel3}
\end{table}

\begin{figure}[!t]
\centering
\includegraphics[width= 0.48\textwidth]{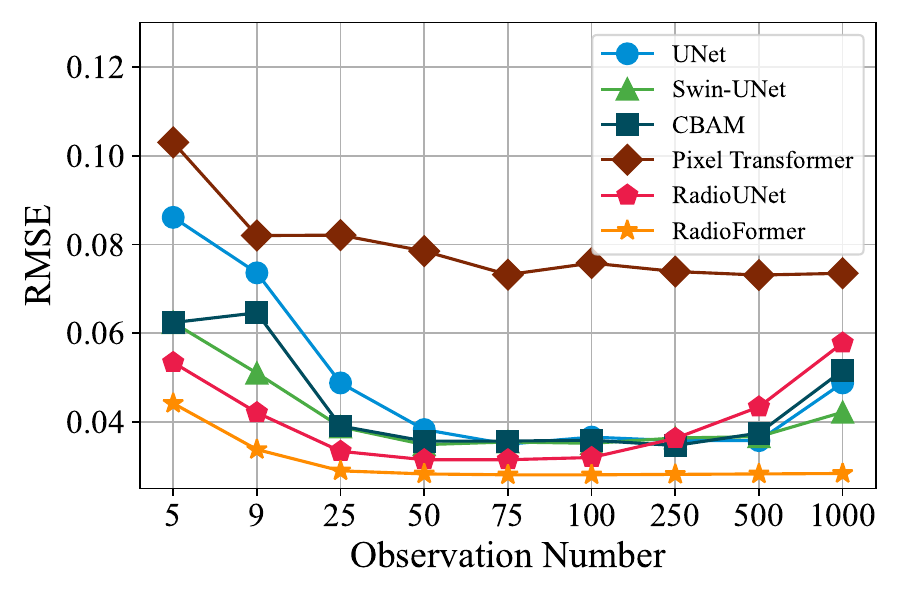}
\caption{The line plot illustrates the relationship between the RMSE~$\downarrow$ values and the number of observation points for the model prediction results.}
\label{fig_5}
\end{figure}

\textbf{Performance on Generalization on Different Number of Observation Points.}
We evaluated the model's generalization ability concerning the number of observation points. Specifically, we used a model trained with 9 points to make predictions across varying observation sizes. 
Fig.~\ref{fig_4} presents the visualized results of the model's predictions for different numbers of observation points, while Tab.~\ref{tabel3} summarizes the model's performance for each observation configuration.

In theory, increasing the number of observation points delivers the model more information for radio map estimation, leading to improved predictions.   
However, our results demonstrate that this relationship is not strictly linear.   
Specifically, performance improves across all models when the number of observation points increases from 9 to 100.   

However, as the number of observation points continues to increase, the performance of certain models begins to deteriorate.   
As shown in Fig.~\ref{fig_4}, when the number of observation points reaches 1000, some models exhibit significant issues in their predictions.   
For instance, CBAM predicts 2 transmitters. UNet, RadioUNet, and Swin-UNet incorrectly predict the location of the signal source or predict some signal noise.   

We present the relationship between RMSE and the number of observation points to further illustrate this performance variation.
As shown in Fig.~\ref{fig_5}, the curves for UNet, Radio-UNet, CBAM, and Swin-UNet exhibit similar patterns, characterized by an initial decline followed by an increase, with Radio-UNet showing the most shift.   
Specifically, when the number of observation points increases from 250 to 1000, All 4 models mentioned above experience a drop in performance.   
In contrast, the curves for Pixel Transformer and RadioFormer display minimal fluctuation, indicating a high level of stability.   
However, the observation that raising the number of sampling points does not lead to improved prediction performance suggests that these models cannot fully exploit the information provided by the additional sampling points.

\begin{table}[!t]
\centering
\caption{Speed Performance}.
\begin{tabular}{@{}l|ccc@{}}
\toprule
                  & Flops      & Parameters & Inference Time \\ \midrule
UNet              & 5.11GFlops & 0.22M      & 0.091s         \\
CBAM              & 5.16GFlops & 0.23M      & 0.151s         \\
Swin-UNet         & 8.08GFlops & 27.2M      & 0.065s        \\
Pixel Transformer & 0.46TFlops & 23.64M     & 29.75s          \\
\rowcolor[HTML]{EFEFEF} 
RadioFormer (ours) & 3.28GFlops & 10.7M      & 0.031s         \\ \bottomrule
\end{tabular}
\label{tabel4}
\end{table}

\textbf{Speed Performance}. Tab.~\ref{tabel4} compares inference volume, model size, and inference speed. 
Our model significantly reduces computational complexity by avoiding the point-by-point reasoning approach compared to the aforementioned pixel-wise transformers. 
Regardless of whether our model is larger or smaller, it consistently achieves the shortest inference time compared to patch-level vision models. 
This efficiency is due to our model's ability to avoid regions with perceptual redundancy, where additional reasoning would be ineffective.

\subsection{Ablation Study}

\begin{table}[t]
\centering
\caption{Ablation Study: Feature Fusion Methods}
\begin{tabular}{@{}l|ccc@{}}
\toprule
\multicolumn{1}{c|}{}                                                                                                       & \multicolumn{3}{c}{Dataset: DPM}                     \\ \cmidrule(l){2-4} 
\multicolumn{1}{c|}{\multirow{-2}{*}{\begin{tabular}[c]{@{}c@{}}Feature Fusion Between\\ Position and Values\end{tabular}}} & RMSE~$\downarrow$         & SSIM~$\uparrow$            & PSNR~$\uparrow$              \\ \midrule
\rowcolor[HTML]{F2F2F2} 
\cellcolor[HTML]{EFEFEF}Cross Attention                                                                                     & \textbf{0.0338} & \textbf{0.9366} & \textbf{29.4259} \\
Channel-wise Self-Attention                                                                                                 & 0.0501          & 0.8691          & 25.9154          \\
Embedding Concation                                                                                                         & 0.0643          & 0.8381          & 23.9207          \\ \midrule
\multicolumn{1}{c|}{}                                                                                                       & \multicolumn{3}{c}{Dataset: DPM}                     \\ \cmidrule(l){2-4} 
\multicolumn{1}{c|}{\multirow{-2}{*}{\begin{tabular}[c]{@{}c@{}}Feature Fusion Between\\ Position and Values\end{tabular}}} & RMSE~$\downarrow$            & SSIM~$\uparrow$             & PSNR~$\uparrow$              \\ \midrule
\rowcolor[HTML]{F2F2F2} 
\cellcolor[HTML]{EFEFEF}Add                                                                                                 & \textbf{0.0338} & \textbf{0.9366} & \textbf{29.4259} \\
Concat                                                                                                                      & 0.0548          & 25.3835         & 0.8532           \\ \midrule
\multicolumn{1}{c|}{}                                                                                                       & \multicolumn{3}{c}{Dataset: DPM}                     \\ \cmidrule(l){2-4} 
\multicolumn{1}{c|}{\multirow{-2}{*}{Position Embedding Methods}}                                                           & RMSE~$\downarrow$            & SSIM~$\uparrow$             & PSNR~$\uparrow$              \\ \midrule
\rowcolor[HTML]{F2F2F2} 
\cellcolor[HTML]{EFEFEF}Sinusoidal                                                                                          & \textbf{0.0338} & \textbf{0.9366} & \textbf{29.4259} \\
Learnable                                                                                                                   & 0.0548          & 0.8532         &  25.3835    \\
Without                                                                                                                     & 0.0663          & 0.8223          & 24.6187          \\ \bottomrule
\end{tabular}
\label{ablation1}
\end{table}

\begin{table}[t]
\centering
\caption{Ablation Study: Embedding Dimensions}
\begin{tabular}{@{}l|c|ccc@{}}
\toprule
\multicolumn{1}{c|}{\multirow{2}{*}{\begin{tabular}[c]{@{}c@{}}Embedding \\ Dimension\end{tabular}}} & \multirow{2}{*}{Parameters} & \multicolumn{3}{c}{Dataset: DPM} \\ \cmidrule(l){3-5} 
\multicolumn{1}{c|}{}                                                                                &                             & RMSE~$\downarrow$      & SSIM~$\uparrow$     & PSNR~$\uparrow$      \\ \midrule
96                                                                                                   & 1.10M                       & 0.0415    & 0.9108   & 27.6342   \\
192                                                                                                  & 3.03M                       & 0.0338    & 0.9366   & 29.4259   \\
384                                                                                                  & 10.42M                      & 0.0421    & 0.9037   & 27.5191   \\
576                                                                                                  & 22.53M                      & 0.0451    & 0.9107   & 26.9236   \\
768                                                                                                  & 39.35M                      & 0.1001    & 0.8838   & 19.992    \\
1024                                                                                                 & 69.08M                      & 0.1009    & 0.8698   & 19.9277   \\ \bottomrule
\end{tabular}
\label{table: embed}
\end{table}

\begin{table}[t!]
\centering
\caption{Ablation Study: The Impact of Reversed Building Map}
\begin{tabular}{@{}l|c|ccc@{}}
\toprule
\multicolumn{1}{c|}{\multirow{2}{*}{}} & \multirow{2}{*}{\begin{tabular}[c]{@{}c@{}}Building map\\ Reverse\end{tabular}} & \multicolumn{3}{c}{Dataset: DPM} \\ \cmidrule(l){3-5} 
\multicolumn{1}{c|}{}                  &                                                                                 & RMSE~$\downarrow$      & SSIM~$\uparrow$     & PSNR~$\uparrow$      \\ \midrule
\multirow{2}{*}{UNet}                  & Yes                                                                             & 0.0736    & 0.8892   & 22.6635   \\
                                       & No                                                                              & 0.8211    & 0.8732   & 22.0076   \\ \midrule
\multirow{2}{*}{CBAM}                  & Yes                                                                             & 0.0523    & 0.9192   & 25.6307   \\
                                       & No                                                                              & 0.0545    & 0.9101   & 25.0641   \\ \midrule
\multirow{2}{*}{Swin-UNet}             & Yes                                                                             & 0.0510    & 0.8927   & 25.8593   \\
                                       & No                                                                              & 0.5321    & 0.8878   & 24.3967   \\ \midrule
\multirow{2}{*}{RadioFormer}           & Yes                                                                             & 0.0388    & 0.9203   & 28.2388   \\
                                       & No                                                                              & 0.0338    & 0.9366   & 29.4259   \\ \bottomrule
\end{tabular}
\label{table: reverse}
\end{table}

To rigorously assess the contribution of each component in RadioFormer to its overall performance, we conducted a series of ablation studies using the DPM dataset. To clarify, all ablation study experiments are set the number of observation points to 9.
First, we evaluated the impact of fusing observation point features with architectural features to determine their role in enhancing model performance. Second, we examined the effectiveness of the feature fusion strategy that integrates the position and value of observation points. Finally, we investigated the influence of the position encoding mechanism applied to the building map’s ViT module on the model’s predictive capabilities. The results of these experiments are summarized in Tab.~\ref{ablation1}, providing a detailed breakdown of each component’s contribution to RadioFormer’s performance.

Additionally, we explored the relationship between the embedding dimension and model performance, as summarized in Tab.~\ref{table: embed}.  
Contrary to expectations based on scaling laws, the model’s performance did not consistently improve with increasing embedding dimensions.  
Instead, we identified an optimal embedding dimension of 192.  Notably, when the embedding dimension exceeded 768, the model’s performance degraded significantly, rendering it ineffective.  
This unexpected behavior underscores the importance of carefully selecting architectural parameters for RadioFormer.

Finally, we explored an intriguing question: Does a reverse building map impact model performance? For baseline models such as U-Net, CBAM, and Swin-UNet, the results shown in Tab.~\ref{table: reverse} consistently indicate that a reverse building map enhances their performance. 
However, the opposite is observed for RadioFormer, where reverse building maps lead to a decline in performance.